\documentclass[letter, 10 pt, journal, twoside]{IEEEtran} 

\usepackage{amsmath}  % assumes amsmath package installed
\usepackage{amssymb}  % assumes amsmath package installed
\usepackage{amsfonts}
\usepackage{mathtools}
\DeclareMathAlphabet{\pazocal}{OMS}{zplm}{m}{n}

\usepackage{csquotes}
\usepackage{paralist}

\usepackage{verbatim}

\usepackage{color}
\usepackage{graphicx}
\usepackage{adjustbox}

\usepackage{mathptmx}
\usepackage{times}
\usepackage[hang,flushmargin]{footmisc}
\usepackage{multirow}

\usepackage[ruled,vlined,linesnumbered]{algorithm2e}
\usepackage[noend]{algorithmic}
\SetKwInput{KwInput}{Input}                % Set the Input
\SetKwInput{KwOutput}{Output}              % set the Output

\usepackage[us]{datetime}
\usepackage{caption}
\newcommand{\tran}{^\top}

\usepackage[normalem]{ulem}

\usepackage[usenames,dvipsnames,table]{xcolor}

\newcommand{\Hl}[2][\empty]{%
\ifx#1\empty
\else
\sethlcolor{#1}%
\fi
\hl{#2}}
\usepackage{soul}
\soulregister\Hl{7}
\soulregister\ref7
\soulregister\autoref7
\soulregister\eqref7
\soulregister\cite7
\soulregister\pageref7
\soulregister\ac7
\soulregister\acp7
\soulregister\acs7
\soulregister\acl7
\soulregister\subsection7
\soulregister\SI7
\soulregister\url7

\newcommand{\rev}[1]{#1} %remove revision
\usepackage{mdframed}

\newmdenv[
   backgroundcolor=yellow,
   linewidth=0pt,
   innertopmargin=1pt,
   innerbottommargin=1pt
]{highlighted}
\newcommand{\cmark}{\color{green}{\ding{51}}}%

\usepackage{hyperref} %pdf with links and toc on the left
\hypersetup{
  colorlinks,
  citecolor=black,
  filecolor=black,
  linkcolor=black,
  urlcolor=black,
  pdfauthor={},
  pdfsubject={},
  pdftitle={}
}

\usepackage{todonotes}

\usepackage{subcaption}

\usepackage{latexsym}
\usepackage{color}

\usepackage{cite}

\usepackage{siunitx}
\usepackage{ifthen}

\usepackage{booktabs}
\usepackage{tablefootnote}

\usepackage{subfiles}
\usepackage{bm}

\usepackage{pifont}
\definecolor{color_green}{rgb}{0, .722, .243}
\definecolor{color_red}{rgb}{0.737,0.165,0}

\newcommand{\xmark}{{\color{color_red}{\ding{55}}}}

\usepackage{stackengine} 

\usepackage{tikz}
\usetikzlibrary{calc}
\usepackage{tikz-dimline}
\usepackage{physics}
\usepackage{tikz-3dplot}
\usepackage[outline]{contour} % glow around text
\usetikzlibrary{fadings}
\usetikzlibrary{decorations.markings}

\tikzfading[name=fade up,top color=red!100, bottom color=white!0]

\tikzset{testfade/.style n args={3}{
    postaction={
    decorate,
    decoration={
    markings,
    mark=between positions 0 and \pgfdecoratedpathlength step 0.5pt with {
    \pgfmathsetmacro\myval{multiply(
        divide(
        \pgfkeysvalueof{/pgf/decoration/mark info/distance from start}, \pgfdecoratedpathlength
        ),
        100
    )};
    \pgfsetfillcolor{#3!\myval!#2};
    \pgfpathcircle{\pgfpointorigin}{#1};
    \pgfusepath{fill};}
}}}}

\colorlet{veccol}{green!45!black}
\colorlet{myred}{red!80!black}
\colorlet{myblue}{blue!80!black}
\colorlet{mypurple}{blue!50!red!100!}
\colorlet{amber}{red!90!yellow!100!blue!50}
\colorlet{mygreen}{green!40!black}
\colorlet{projcol}{blue!70!black}
\colorlet{mydarkblue}{blue!50!black}
\colorlet{veccol}{green!50!black}
\usetikzlibrary{shapes.geometric,arrows.meta,angles,quotes}
\tikzstyle{thin arrow}=[dashed,thin,-{Latex[length=4,width=3]}]
\tikzstyle{arrow}=[{-Latex}]

\tikzset{>=latex} % for LaTeX arrow head
\tikzstyle{proj}=[projcol!80,line width=0.08] %very thin
\tikzstyle{area}=[draw=veccol,fill=veccol!80,fill opacity=0.6]
\tikzstyle{vector}=[-stealth,myblue,thick,line cap=round]
\tikzstyle{unit vector}=[->,veccol,thick,line cap=round]
\tikzstyle{dark unit vector}=[unit vector,veccol!70!black]
\usetikzlibrary{angles,quotes} % for pic (angle labels)
\contourlength{1.3pt}

\definecolor{color_blue}{rgb}{0.22, 0.2, 0.502}
\definecolor{color_red}{rgb}{0.737,0.165,0}
\definecolor{color_green}{rgb}{0, .522, .243}

\usepgfplotslibrary{fillbetween}
\usepgfplotslibrary{groupplots}
\def\centerarc[#1](#2)(#3:#4:#5)% Syntax: [draw options] (center) (initial angle:final angle:radius)
    { \draw[#1] ($(#2)+({#5*cos(#3)},{#5*sin(#3)})$) arc (#3:#4:#5); }

\DeclareMathOperator{\diag}{diag}

\usepackage{calrsfs}
\DeclareMathAlphabet{\pazocal}{OMS}{zplm}{m}{n}

\title{%
\Title
}

\usepackage[printonlyused]{acronym}
    \acrodef{acw}[ACW]{Aerial Co-Worker}
    \acrodef{uav}[UAV]{Unmanned Aerial Vehicle}
    \acrodef{ar}[AR]{Aerial Robot}
    \acrodef{dnn}[DNN]{Deep Neural Network}
    \acrodef{fov}[FoV]{Field of View}
    \acrodef{hsi}[HSI]{Human-Swarm Interaction}
    \acrodef{lstm}[LSTM]{Long Short-Term Memory Network} 
    \acrodef{mrav}[MRAV]{Multi-Rotor Aerial Vehicle}
    \acrodef{nmpc}[NMPC]{Nonlinear Model Predictive Control}
    \acrodef{ssd}[SSD]{Single-Shot multibox Detector}
    \acrodef{uwb}[UWB]{Ultra Wide Bandwidth}
    \acrodef{cnn}[CNN]{Convolutional Neural Network}
    \acrodef{lstm}[LSTM]{Long Short-Term Memory}
    \acrodef{fcu}[FCU]{Flight Control Unit}
    \acrodef{imu}[IMU]{Inertial Measurement Unit}
    \acrodef{ros}[ROS]{Robot Operating System}
    \acrodef{wrt}[w.r.t.]{with respect to}
\title{Gesture-Controlled Aerial Robot Formation for Human-Swarm Interaction in Safety Monitoring Applications}

\author{V\'{i}t Kr\'{a}tk\'{y}$^{1\star}$, Giuseppe Silano$^{1,2}$, Matou\v{s} Vrba$^1$, Christos Papaioannidis$^3$, Ioannis Mademlis$^3$, Robert P\v{e}ni\v{c}ka$^1$,\\ Ioannis Pitas$^3$, and Martin Saska$^1$
  \thanks{Manuscript received: September, 10, 2024; Revised March, 25, 2025; Accepted June, 7, 2025.}
  \thanks{This paper was recommended for publication by Editor Venture Gentiane upon evaluation of the Associate Editor and Reviewers' comments. This work was partially funded by the EU's H2020 AERIAL-CORE grant no. 871479, by the CTU grant no. SGS23/177/OHK3/3T/13, by the GAČR grant no. 23-07517S, by the  EU under ROBOPROX reg. no. CZ.02.01.01/00/22\_008/0004590, and by the research fund for the Italian Electrical System under the decree n. 388 of November 6th, 2024.}
    \thanks{$^1$Authors are with Department of Cybernetics, Faculty of Electrical Engineering, Czech Technical University in Prague, Czech Republic {\tt\footnotesize\{\href{mailto:vit.kratky@fel.cvut.cz}{vit.kratky}|\href{mailto:giuseppe.silano@fel.cvut.cz}{giuseppe.silano}|\href{mailto:matous.vrba@fel.cvut.cz}{matous.vrba}|\\\href{mailto:robert.penicka@fel.cvut.cz}{robert.penicka}|\href{mailto:martin.saska@fel.cvut.cz}{martin.saska}\}@fel.cvut.cz}, $^2$Author is with Department of Power Generation Technologies and Materials, Ricerca sul Sistema Energetico, Italy. $^3$Authors are with Department of Informatics, Aristotle University of Thessaloniki, Greece {\tt\footnotesize\{\href{mailto:cpapaionn@csd.auth.gr}{cpapaionn}|\href{mailto:imademlis@csd.auth.gr}{imademlis}|\href{mailto:pitas@csd.auth.gr}{pitas}\}@csd.auth.gr}. $^\star$Corresponding author.}
\thanks{Digital Object Identifier (DOI): see top of this page.}

}%
\newcommand{\PREPRINTYEAR}{2025}
\newcommand{\PUBLISHEDIN}{IEEE Robotics and Automation Letters }
\newcommand{\DOI}{10.1109/LRA.2025.3583607} % you will not get a DOI until the paper is actually published, so update this when you get it and reupload the new preprint to all systems

\usepackage[placement=top,vshift=-2,firstpage=true]{background}
\SetBgScale{1.0}
\SetBgContents{\parbox{0.95\textwidth}{\small \begin{center} The manuscript is published in \PUBLISHEDIN under DOI: \DOI\end{center} \vspace{-0.2cm}  \copyright{}~\PREPRINTYEAR~IEEE.~Personal use of this material is permitted. Permission from IEEE must be obtained for all other uses, in any current or future media, including reprinting/republishing this material for advertising or promotional purposes, creating new collective works, for resale or redistribution to servers or lists, or reuse of any copyrighted component of this work in other works.} }
\SetBgColor{black}
\SetBgAngle{0}
\SetBgOpacity{1.0}

\begin{document}
\maketitle
\setcounter{footnote}{3}
\linepenalty=1000  % Increase to discourage single-word lines

%%% START SECTION ==========================================================

% %%{ Abstract
\begin{abstract}

 This paper presents a formation control approach for contactless gesture-based \ac{hsi} between a team of multi-rotor~\acp{uav} and a human worker. 
 The approach is designed to monitor the safety of human workers, particularly those operating at heights. 
 In the proposed dynamic formation scheme, one~\ac{uav} acts as the formation leader, equipped with sensors for detecting human workers and recognizing gestures. 
 The follower~\acp{uav} maintain a predetermined formation relative to the worker's position, providing additional perspectives of the monitored scene. 
 Hand gestures enable the human worker to specify movement and action commands for the~\ac{uav} team and to initiate other mission-related tasks without requiring additional communication channels or specific markers.
 Combined with a novel unified human detection and tracking algorithm, a human position estimation method, and a gesture detection pipeline, the proposed approach represents the first instance of an \ac{hsi} system incorporating all these modules onboard real-world \acp{uav}.
 Simulations and field experiments involving three~\acp{uav} and a human worker in a mock-up scenario demonstrate the effectiveness and responsiveness of the proposed approach.
  
\end{abstract}

%%}

%%% END SECTION ============================================================

%%% START SECTION ==========================================================

%%{ Keywords

%%}

%%% END SECTION ============================================================

%%% START SECTION ==========================================================
% \vspace{-0.2cm}
\section*{Supplementary material}
% \vspace{-0.2cm}
\textbf{Video:} {\small\url{https://mrs.felk.cvut.cz/gestures2024}} 

%%% END SECTION ============================================================

%%% START SECTION ==========================================================

%%{ SECTION: INTRODUCTION

\section{Introduction}
\label{sec:introduction}

\IEEEPARstart{T}{he} application of multi-rotor~\acfp{uav} in challenging-to-access real-world environments, such as wind turbines~\cite{MANSOURI2018118}, large construction sites~\cite{Loianno2018IJRR}, and power transmission lines~\cite{CaballeroIEEEAccess2023}, has proven to be exceptionally beneficial.

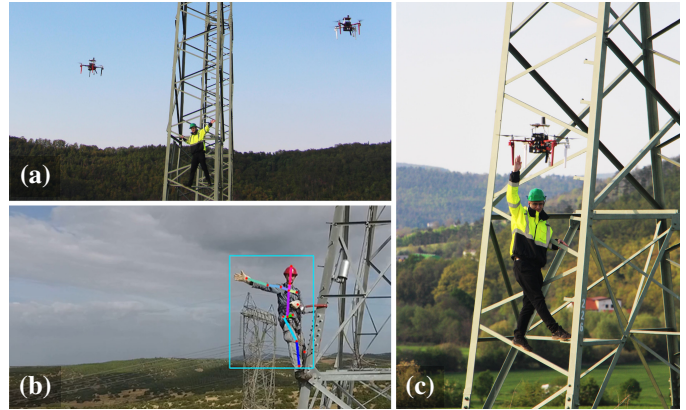
\begin{figure}[tb]
  \centering
  \begin{tikzpicture}
    \node[anchor=south west,inner sep=0] (a) at (0,0) {
      \includegraphics[width=1.00\columnwidth]{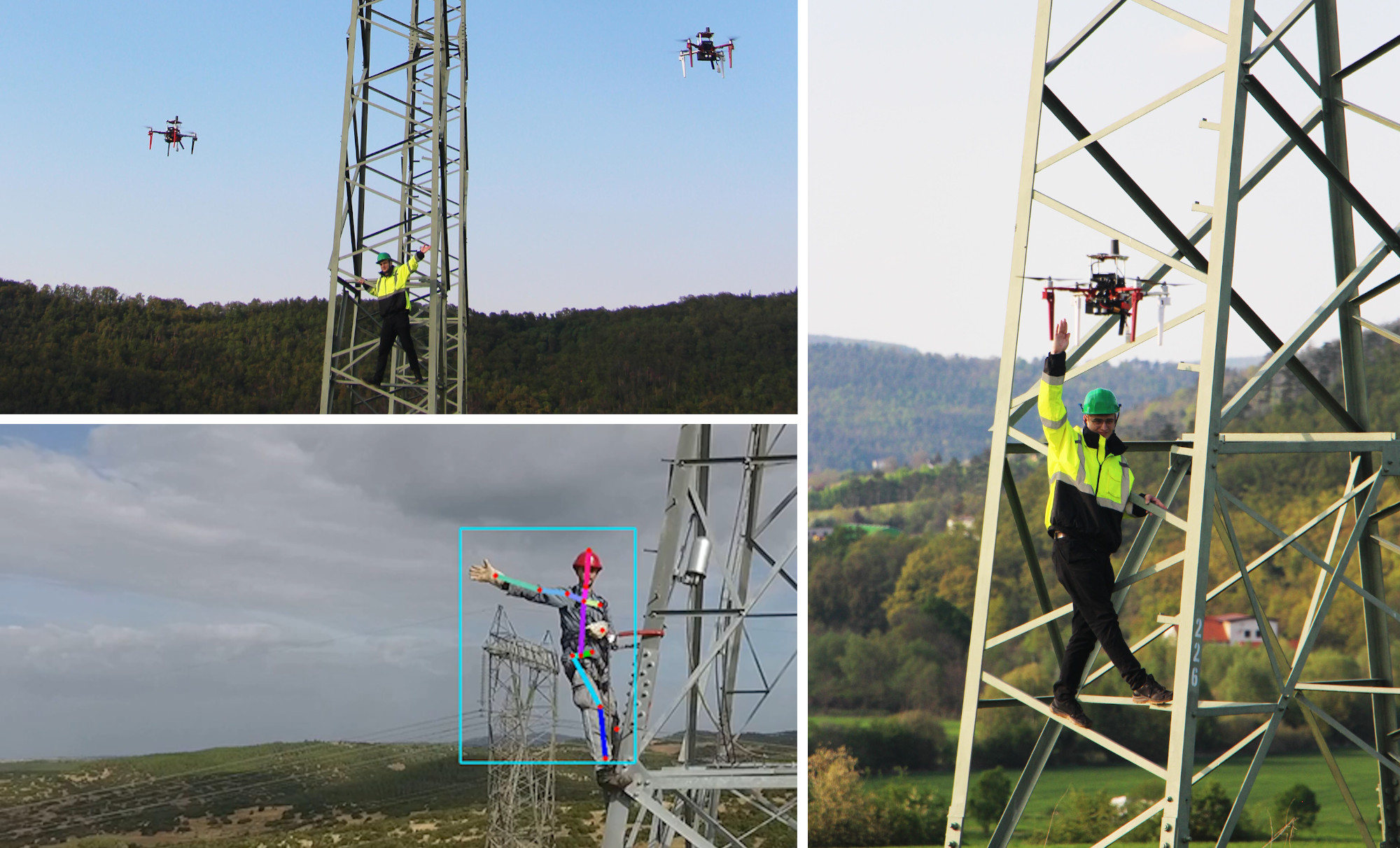}
    };

    \pgfmathsetmacro{\hzero}{0.039}
    \pgfmathsetmacro{\vzero}{0.5708}
    \pgfmathsetmacro{\hshift}{0.5738}
    \pgfmathsetmacro{\vshift}{-0.5175}
   
    \begin{scope}[x={(a.south east)},y={(a.north west)}]

        \node[fill=black, fill opacity=0.3, text=white, text opacity=1.0] at (\hzero, \vzero) {\textbf{(a)}};
        \node[fill=black, fill opacity=0.3, text=white, text opacity=1.0] at (\hzero, \vzero+\vshift) {\textbf{(b)}};
        \node[fill=black, fill opacity=0.3, text=white, text opacity=1.0] at (\hzero+\hshift, \vzero+\vshift) {\textbf{(c)}};

    \end{scope}

  \end{tikzpicture}
  % \vspace*{-1.5em}
  \caption{The gesture-based interaction between a human worker and a team of \acp{uav} using the proposed system (a, c). An example output from the developed gesture recognition pipeline is overlaid on the corresponding input video frame (b).}
  \label{fig:safetyACWdemo}
  % \vspace*{-1.9em}
\end{figure}
\acp{uav}, serving as \textit{robotic co-workers}~\cite{Haddadin2011Springer} in these settings, offer numerous advantages, including the ability to access hard-to-reach locations, assist in tool handling, monitor worker safety, and reduce both the physical and cognitive workloads of human workers~\cite{TognonTRO2021, BenziRAL2022}. 
Within the framework of the European AERIAL-CORE project\footnote{\url{https://aerial-core.eu}}, the emphasis on safety monitoring stems from the observation that violations of safety protocols are a leading cause of fatal injuries during maintenance tasks on electric power infrastructures.
To address this issue, the concept of \textit{Aerial Co-Workers} (\acsp{acw}) has been developed~\cite{olleroAERAM}, encompassing three roles: the inspection-\acs{acw} \cite{CaballeroIEEEAccess2023}, the safety-\acs{acw} \cite{kratky2021aerialfilming}, and the physical-\acs{acw} \cite{Afifi2022ICRA}. These roles are envisioned as key components of future human-robot missions aimed at maintaining electric power transmission infrastructures and, more broadly, the entire energy system. %, including offshore wind turbines, solar farms, and other energy-generating facilities.  

In safety monitoring applications, it is crucial that the human operator responsible for situation assessment is provided with a comprehensive view of the scene.
The comprehensive view is greatly enhanced by the ability to adapt the perspective interactively, with situational awareness significantly improving as the number of simultaneous scene perspectives increases.
This need underscores the utility of deploying multiple \acp{uav} for safety monitoring. However, when monitoring human workers, it is essential to balance the operator's preferences with the safety and comfort of the monitored individuals, ensuring that their performance is not adversely affected by the proximity of the \acp{uav}.

\begin{table*}[tb]
    \setlength{\tabcolsep}{-1.0pt}
    \centering
    \renewcommand{\arraystretch}{0.97}
    \vspace*{0.13cm}
    \caption{Comparison of the addressed features in related papers and our proposed approach: \textit{included} ({\cmark}) and \textit{not included} ({\xmark}).}
    \label{tab:comparison}
    \vspace{-0.65em}
    % \scalebox{1.02}{
    \begin{tabular}{@{}l c c c c c c c c c c c @{}}
    \hline
      \multirow{2}{*}{\textbf{Reference~~}} & \multicolumn{7}{c}{\textbf{Features}} & \phantom{x} & \multicolumn{3}{c}{\textbf{Evaluation}} \\ \cline{2-8} \cline{10-12} \rule{0pt}{1ex} 
  & $\begin{array}{cc}\textbf{Human}\\\textbf{Gestures}\end{array}$ & $\begin{array}{cc}\textbf{Multi-robot}\\\textbf{Control}\end{array}$ & $\begin{array}{cc} \textbf{Onboard}\\\textbf{Computation}\end{array}$ & $\begin{array}{cc}\textbf{Adaptive}\\\textbf{Parameters}\end{array}$ & $\begin{array}{cc}\textbf{Human}\\\textbf{Tracking}\end{array}$ & $\begin{array}{cc}\textbf{Obstacle}\\\textbf{Avoidance}\end{array}$ &
        $\begin{array}{cc}\textbf{Mutual Collision}\\\textbf{Avoidance}\end{array}$ & \phantom{x} & $\begin{array}{c}\textbf{Simulation}\end{array}$ & $\begin{array}{c}\textbf{Lab. Env.}\end{array}$ & $\begin{array}{c}\textbf{Outdoors}\end{array}$ \\ %\hline
     \cite{Nagi2014IROS} & {\cmark} & {\cmark} & {\xmark} & {\cmark} & {\xmark} & {\xmark} & {\xmark} & & {\xmark} & {\cmark} & {\xmark} \\ %\hline
      \cite{Abbate2021ICRA} & {\cmark} & {\xmark} & {\xmark} & {\xmark} & {\xmark} & {\xmark} & {\xmark} & & {\xmark} & {\cmark} & {\xmark} \\ % \hline  
    % \cite{CoutureCCRV2010} & {\xmark} & {\xmark} & {\xmark} & {\cmark} & {\xmark} & {\xmark} \\ % \hline
      \cite{Abdi2023ICRA} & {\cmark} & {\cmark} & {\xmark} & {\cmark} & {\xmark} & {\cmark} & {\cmark} & & {\xmark} & {\cmark} & {\xmark} \\ % \hline 
      \cite{MonajjemiIROS2013} & {\cmark} & {\cmark} & {\xmark} & {\xmark} & {\xmark} & {\xmark} & {\xmark} & & {\xmark} & {\cmark} & {\xmark} \\ %\hline
      \cite{Jacquet2022RAL} & {\xmark} & {\cmark} & {\cmark} & {\xmark} & {\xmark} & {\xmark} & {\cmark} & & {\cmark} & {\cmark} & {\xmark} \\ % \hline
      \cite{akash2022icuas} & {\cmark} & {\cmark} & {\cmark} & {\xmark} & {\xmark} & {\xmark} & {\cmark} & & {\cmark} & {\xmark} & {\cmark} \\ %\hline
    % \cite{AlcantaraRAS2021} & {\xmark} & {\cmark} & {\cmark} & {\xmark} & {\xmark} & {\cmark} & {\cmark} & & {\cmark} & {\xmark} & {\cmark} \\ % \hline
    % \cite{Tallamraju2019RAL} & {\xmark} & {\cmark} & {\cmark} & {\xmark} & {\cmark} & {\cmark} & {\cmark} & & {\cmark} & {\xmark} & {\cmark} \\ % \hline
      \cite{Price2018RAL} & {\xmark} & {\cmark} & {\cmark} & {\xmark} & {\cmark} & {\xmark} & {\cmark} & & {\xmark} & {\xmark} & {\cmark} \\ % \hline
      \cite{XuTVCG2018} & {\xmark} & {\cmark} & {\cmark} & {\xmark} & {\cmark} & {\xmark} & {\cmark} & & {\xmark} & {\cmark} & {\cmark} \\ % \hline
      \cite{kratky2021aerialfilming, AlcantaraRAS2021} & {\xmark} & {\cmark} & {\cmark} & {\xmark} & {\xmark} & {\cmark} & {\cmark} & & {\cmark} & {\xmark} & {\cmark} \\ % \hline
    % \cite{Saini2019ICCV}  & {\xmark} & {\cmark} & {\cmark} & {\xmark} & {\xmark} & & & & &  \\ % similar to tallamraju2019ral
      \cite{HoIROS2021, Tallamraju2019RAL}  & {\xmark} & {\cmark} & {\cmark} & {\xmark} & {\cmark} & {\cmark} & {\cmark} & & {\cmark} & {\xmark} & {\cmark} \\ % \hline
      ~Ours & {\cmark} & {\cmark} & {\cmark} & {\cmark} & {\cmark} & {\cmark} & {\cmark} & & {\cmark} & {\xmark} & {\cmark} \\ \hline
    \end{tabular}
    % }
    \vspace{-0.6cm}
\end{table*}

This study introduces a novel approach for \ac{uav} formation control in contactless~\acf{hsi}, focusing on teams of multi-rotor \acp{uav}. 
By leveraging gesture-based controls, our approach aims to enhance situational awareness and facilitate precise command execution in real-world scenarios, such as maintenance operations on electric power transmission infrastructures. 
The framework enables a remote operator to dynamically adjust the \ac{uav} formation to optimize observation angles, while also allowing a monitored worker to use gestures to request assistance, modify the \acp{uav}' proximity or abort the mission for safety reasons. 
This dual-control mechanism ensures the system can adapt in real time to the operator's needs and the task context, while empowering the monitored individual to influence the \acp{uav}' behavior without requiring additional equipment, such as wearable sensors. 
This capability is crucial for ensuring worker safety during unforeseen events, as it leverages the worker's superior awareness of \ac{uav} proximity, nearby obstacles, and prevailing weather conditions compared to the remote operator. 

The work presented in this paper bridges the gap in human-multi \ac{uav} interaction by introducing an innovative system that integrates advanced \ac{hsi} features, combining vision and control strategies directly onboard \ac{uav} platforms for seamless and responsive collaboration. 
Validated through rigorous testing in both simulated and real-world outdoor conditions, as illustrated in \autoref{fig:safetyACWdemo} and~\autoref{fig:simulation}, the proposed system demonstrates significant advancements in practical deployment, and a potential to enhance safety and efficiency in critical infrastructure maintenance and inspection tasks.

%%% END SECTION ============================================================

%%% START SECTION ==========================================================

\subsection{Related Work}
\label{sec:relatedWork}

Extensive research has been conducted on collaborative and safe interactions involving human and ground robots; however, methods involving \acp{uav} in this context remain less developed~\cite{Ajoudani2018AR}. In particular, the dynamics of human interaction with multi-robot \ac{uav} teams present a significant research gap. 
While there has been considerable advancement in computer vision and autonomous systems to facilitate human-\ac{uav} interaction, these efforts are often limited to specific sub-problems~\cite{KollingHMS2016, DahiyaRAS2023}. 
For example, studies in computer vision have focused on recognizing human features such as faces~\cite{CoutureCCRV2010}, hand gestures~\cite{Nagi2014IROS, Abbate2021ICRA, Abdi2023ICRA}, hand motion~\cite{Macchini2021RAL}, and body postures~\cite{MonajjemiIROS2013}, with some
exploring gaze detection for selecting robots in multi agent scenarios~\cite{zhang2016gazeSelection}. 
Meanwhile, research on \ac{uav} autonomy has addressed perception-aware control~\cite{Jacquet2022RAL}, formation control to enhance visibility~\cite{kratky2021aerialfilming}, and optimization-based obstacle avoidance~\cite{AlcantaraRAS2021}, all aimed at improving \acp{uav}' independent navigation and safety around humans and other \acp{uav}.

Despite these advances, integrating these technologies into cohesive systems for human-\ac{uav} teams operating in complex, real-world environments remains underexplored~\cite{BonattiJFR2020}.
Current studies often focus heavily on the vision component, sometimes neglecting or oversimplifying the dynamics of \acp{uav} or introducing unrealistic assumptions on the environment~\cite{CoutureCCRV2010, Nagi2014IROS, Abbate2021ICRA, Abdi2023ICRA, akash2022icuas}, or they focus on control aspects without adequately leveraging onboard sensor data~\cite{Jacquet2022RAL, kratky2021aerialfilming, AlcantaraRAS2021}.
This lack of a holistic approach in designing~\ac{hsi} frameworks can lead to critical failures.
For instance, inaccuracies in estimating human position due to factors like unbalanced camera vibration or motion blur can compromise system stability, potentially resulting in crashes and endangering the operator.
Moreover, most previous studies~\cite{Nagi2014IROS, Abdi2023ICRA, Macchini2021RAL, MonajjemiIROS2013, XuTVCG2018, HoIROS2021} have not adequately addressed the challenge of integrating onboard gesture recognition modules within the~\ac{uav} formation control schemes. % to enhance~\ac{hsi}, while also ensuring swift responsiveness.
Additionally, many of these methods rely on offboard computation~\cite{Nagi2014IROS, Abbate2021ICRA, Abdi2023ICRA, Macchini2021RAL, CoutureCCRV2010, MonajjemiIROS2013} and have been evaluated only in indoor environments \cite{Nagi2014IROS, Abbate2021ICRA, Abdi2023ICRA, Jacquet2022RAL, Macchini2021RAL, CoutureCCRV2010, MonajjemiIROS2013}, often without considering external factors such as variable lighting conditions, non-ideal self-localization, and wind gusts.

Proactive motion capture with \acp{uav} has been explored in some studies~\cite{XuTVCG2018, Saini2019ICCV, Price2018RAL}, but these works often lack robust obstacle avoidance strategies, limiting their effectiveness in complex environments. % For example, systems like AirCap demonstrate the potential for fully autonomous human motion capture using multiple \acp{uav}, though they still face challenges in matching the accuracy of commercial marker-based systems.
Recent advancements in 3D human pose reconstruction have begun to address these challenges~\cite{Tallamraju2019RAL, HoIROS2021}. However, these works primarily focus on optimizing viewpoint configurations for human reconstruction and neglect aspects of the \ac{hsi} and the corresponding features essential for enabling intuitive and responsive human-\ac{uav} interaction in real-time scenarios.

Building upon our prior work~\cite{Symeonidis2023TIP, karakostas2021occlusion, Papaioannidis2022tcsvt, papaioannidis2021EUSIPCO}, this paper advances beyond the existing methodologies by designing algorithms that jointly address human detection, pose estimation, gesture recognition, and \ac{uav} formation control, all executable onboard lightweight \acp{uav} as a unified system.
This approach eliminates off-board processing latency and enables real-world deployment without dependence on external infrastructure.
The proposed~\ac{hsi} framework also incorporates all relevant safety features such as obstacle avoidance, collision prevention with other~\acp{uav}, and adherence to distance regulations to ensure human comfort and safety. % during interactions with aerial robots.
For a comprehensive comparison of the features addressed in the related works and in the proposed approach, we refer the reader to~\autoref{tab:comparison}.

%%% END SECTION ============================================================

\begin{figure*}[htb]
  \centering
  \input{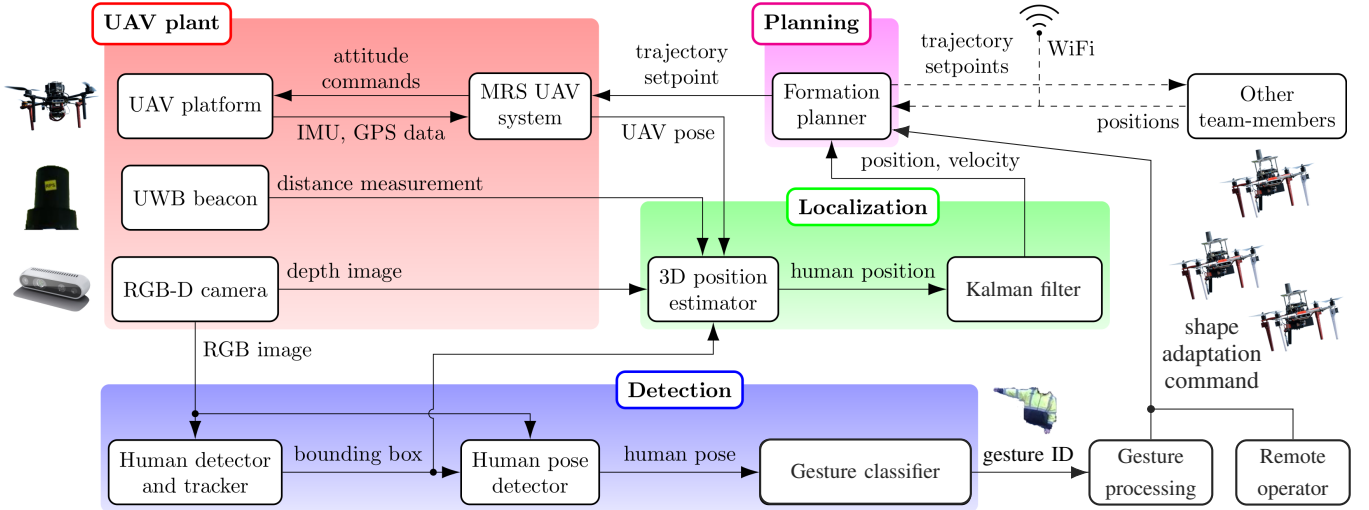}
  \vspace*{-0.1em}
  \caption{A system architecture overview showing data exchange between blocks using arrows and highlighted layers.}
  \vspace*{-1.4em}
  \label{fig:softwareArchitecture}
\end{figure*}

%%% START SECTION ==========================================================

\subsection{Contributions}
\label{sec:contributions}

%\todoin{Add and emphasize the contributions with respect to our previous works.}

This work presents a gesture-based \ac{hsi} framework for \ac{uav} formations featuring fully onboard processing and real-time adaptability. Building on prior work, it introduces the following key contributions:
i) Dynamic formation control strategy that supports online, on-demand adaptation of the \ac{uav} formation shape in complex environments allowing for rapid response to environmental changes and providing an effective method for controlling relative positions of multiple \acp{uav} around a human worker through operator commands;
ii) A gesture-based control interface for markerless \ac{hsi} extending\cite{Symeonidis2023TIP, papaioannidis2021EUSIPCO, Papaioannidis2022tcsvt}, by integrating an enhanced 2D pose estimation network using cross-attention mechanisms, a unified onboard detection-tracking-skeleton-classification pipeline, and a FIFO buffering mechanism for robust temporal input to the \ac{lstm} gesture classifier.
iii) A multi-modal approach to human position estimation specifically tailored for dynamic \ac{uav} systems operating independently of external infrastructure which can work without additional equipment.
Finally, we demonstrate that advanced \ac{hsi} capabilities combining vision-based perception and control can be effectively deployed onboard lightweight \ac{uav} platforms.

%%}

%%% END SECTION ============================================================

%%% START SECTION ==========================================================

%%{ SECTION: Gesture-controlled aerial formation

\section{Gesture-controlled Aerial Formation}
\label{sec:softwareFramework}

The system architecture, as depicted in~\autoref{fig:softwareArchitecture}, consists of four layers: \textit{Detection}, \textit{Localization}, \textit{Planning}, and \textit{\ac{uav} Plant}. The \textit{Detection} layer interfaces directly with the human worker, translating hand gestures into commands for the \ac{uav} formation. An RGB-D camera is used to capture images, which are then processed for human detection, tracking, and gesture recognition (\autoref{subsec:targetDetection}). The \textit{Localization} layer combines sensor data from the~\ac{uav} plant, including the vehicle's relative distance from the worker, with data from the \textit{Detection} layer and an~\ac{uwb} module. This combined  information serves as input to the Kalman filter, which estimates the human's 3D position and velocity for the formation controller (\autoref{sec:human_position_estimation}). The \textit{Planning} layer generates feasible trajectories for the individual \acp{uav} based on several inputs: the current state of the human worker, requests from a remote human operator, outputs from the gesture classifier, and the statuses of other~\ac{uav} team members obtained through a wireless network (\autoref{sec:formation_control}). Lastly, the \textit{\ac{uav} Plant} layer receives the planned trajectories and executes them. This layer manages the low-level control of the \acp{uav}, translating high-level commands into real-world actions to maintain formation and perform dynamic adjustment as needed~\cite{Baca2020mrs}. % and employs a low-level controller~\cite{Baca2020mrs} to track and execute them for precise flight.

%%}

%%% END SECTION ============================================================

%%% START SECTION ==========================================================

%%{ SECTION: Human detection and gesture recognition

\subsection{Human detection and gesture recognition}
\label{subsec:targetDetection}

RGB images from the onboard camera are processed during flight to detect and track the human worker, leveraging the authors' prior work~\cite{Symeonidis2023TIP}. A fast \ac{dnn} object detector based on the \ac{ssd}~\cite{liu2016ssd} architecture is employed alongside a custom LDES-ODDA visual tracker~\cite{karakostas2021occlusion}. \rev{These components are integrated into a unified onboard detection-and-tracking configuration where detection and tracking alternate, enabling high accuracy and fast inference for real-time gesture recognition on embedded hardware.}
The output of this pipeline is a predicted bounding box for the tracked human in each input image where the human is visible, as illustrated in~\autoref{fig:safetyACWdemo}(b). These bounding boxes are subsequently used for gesture recognition and human state estimation. 
To maximize detection accuracy, both the detector and the tracker were pretrained on a manually annotated dataset\footnote{\url{https://aiia.csd.auth.gr/open-multidrone-datasets}} and then fine-tuned using videos of a human operator wearing safety equipment. These videos were captured in diverse environments and under varying lighting conditions to ensure robustness across scenarios.

Given a sequence of images captured by the RGB-D camera of the leader~\ac{uav} and the corresponding bounding boxes of the tracked human, the developed gesture recognition module predicts the type of the gesture from a predefined set (e.g., extending one arm to the side)~\cite{Patrona2021ICIAfS, Patrona2021AIRPHARO}.
The gesture recognition process follows a sequential pipeline. First, the video frame is cropped using the corresponding bounding box of the tracked human. 
The cropped image is processed by an improved version of our earlier multi-branch \ac{cnn} model~\cite{Papaioannidis2022tcsvt}, enhanced with cross-attention synapses \cite{vaswani2017attention} replacing simpler inter-branch skip connections.

The last $N$ outputs of the skeleton extractor, covering $N$ successive video frames, are stored in a FIFO buffer, which \rev{ensures the \ac{lstm} classifier receives temporally consistent input for reliable gesture recognition in dynamic conditions.} This buffer is then processed by our gesture classifier~\cite{papaioannidis2021EUSIPCO}, a lightweight~\ac{lstm} neural architecture that determines the type of performed gesture based on a temporal sliding window of size $N$. The gesture recognition pipeline was trained on a large, manually annotated dataset of gestures\footnote{\url{https://aiia.csd.auth.gr/auth-uav-gesture-dataset}}, and further fine-tuned to perform effectively on aerial images.
In this work we consider four gestures: crossing arms, extending an arm to the side, placing palms together and raising an arm upwards (see~\autoref{fig:system_demo} for illustration).

The output of the gesture recognition pipeline undergoes a post-processing step to enhance \ac{hsi} reliability by mitigating false positive gesture detections. In each iteration, the $K$ most recent valid measurements are considered, with older data filtered out beyond a predefined age threshold $t_c \in \mathbb{R}_{>0}$ to maintain relevance of the data. \rev{Based on the filtered set of measurements $\mathbb{G}$, the dominant gesture's ratio is determined as $f_d = \frac{g_{f,max}}{|\mathbb{G}|} \in [0, 1]$, with $g_{f, max}$ being the maximum number of detections associated with a single gesture in $\mathbb{G}$}. If this ratio exceeds a predefined threshold $\Pi_d \in [0, 1]$, \rev{the formation parameter $q$ mapped to the dominant gesture is adjusted by a predefined increment $i_q \in \mathbb{R}$} (see \autoref{sec:formation_control}). However, to prevent repeated adjustments based on the same set of measurements, a new adjustment can only occur after a time delay $t_d \in \mathbb{R}_{>0}$. This mechanism improves the worker's control and prevents unwanted shape adaptations. The values of $t_c$, $\Pi_d$, and $t_d$ were determined through real-world experiments.
The mapping of available gestures to formation parameters is not fixed and can be adjusted based on the target application.

%%}

%%% END SECTION ============================================================

%%% START SECTION ==========================================================

%%{ SECTION: Human 3D position estimation

\subsection{Human 3D position estimation}
\label{sec:human_position_estimation}

The estimated human's 3D position, denoted as ${^H}\mathbf{p} = [{^H}p_x, {^H}p_y, {^H}p_z]\tran \in \mathbb{R}^3$, is derived from detections and available onboard sensors and is subsequently refined using a Kalman filter. We employ a constant velocity model within the Kalman filter, formulated as:
\begin{align}
  \begin{bmatrix}
    {^H}\mathbf{p} \\
    {^H}\mathbf{v}
  \end{bmatrix}_{[k+1]}
   &=
   \begin{bmatrix}
     \mathbf{I}_3 & \Delta t \mathbf{I}_3 \\
     \mathbf{0}_3 & \mathbf{I}_3
  \end{bmatrix}
  \begin{bmatrix}
    {^H}\mathbf{p} \\
    {^H}\mathbf{v}
  \end{bmatrix}_{[k]}
  + \bm{\varepsilon}_{[k]}, \\
  \mathbf{z}_{[k]} &=
  {^H}\mathbf{p}_{[k]} + \bm{\zeta}_{[k]},
\end{align}
\vspace*{-0.5cm}
\begin{align}
  \bm{\varepsilon}_{[k]} \sim \pazocal{N}\left( \mathbf{0}, \mathbf{Q} \right),
  && \bm{\zeta}_{[k]} \sim \pazocal{N}\left( \mathbf{0}, \bm{\Sigma}_{[k]} \right).
\end{align}
Here, the subscript $\bullet_{[k]}$ indicates the time step, ${^H}\mathbf{v} = [{^H}v_x, {^H}v_y, {^H}v_z]^\top \in \mathbb{R}^3$ represents the human's velocity, $\mathbf{I}_3 \in \mathbb{R}^{3 \times 3}$ is the identity matrix, $\mathbf{0}_3 \in \mathbb{R}^{3 \times 3}$ is the zero matrix, and $\Delta t$ is the time step duration. The vector $\mathbf{z}$ represents the measurement, while $\bm{\varepsilon}$ and  $\bm{\zeta}$ denote the process noise and measurement noise, respectively, both modeled as zero-mean normal distribution. The covariance matrices for these distributions are $\mathbf{Q}$ for the process noise and $\bm{\Sigma}$ for the measurement noise. %, and $\bm{\zeta}$ is the measurement noise with a covariance matrix $\bm{\Sigma}$.
The process noise covariance matrix $\mathbf{Q}$ is defined as:
\begin{equation}
  \mathbf{Q} = \diag\left( \sigma_{p_x}^2, \sigma_{p_y}^2, \sigma_{p_z}^2, \sigma_{v_x}^2, \sigma_{v_y}^2, \sigma_{v_z}^2 \right),
\end{equation}
where $\sigma_{p_\bullet}$ and $\sigma_{v_\bullet}$ are empirically derived parameters. To simplify our notation, we will omit explicit  time dependencies $\bullet_{[k]}$ from now on.

The measurement vector $\mathbf{z}$ is obtained using a unit vector $\vec{\mathbf{d}}$, which indicates the direction from the camera to the human, and a distance estimate ${^C}d$. This direction vector $\vec{\mathbf{d}}$ is determined by projecting the center of the bounding box using a calibrated camera projection model.
The distance ${^C}d$ is computed by \rev{selecting} estimates from three sources:
\begin{enumerate}
  \item \textit{Apparent distance} $d_\mathrm{apparent}$ computed based on the apparent size of the human in the image and the known physical height, using techniques from literature~\cite{vrba2020ral}.
  \item \textit{Stereo camera distance} $d_\mathrm{stereo}$ derived from the median of distance measurements within the bounding box captured by the stereo camera.
  \item \textit{\ac{uwb} system distance} $d_\mathrm{UWB}$ obtained from an \ac{uwb} system\footnote{\url{https://github.com/Terabee/positioning_systems_api}} mounted on the \ac{uav} and worn by the human.
\end{enumerate}

Based on the manufacturers' specifications, $d_\mathrm{UWB}$ is generally considered more accurate than $d_\mathrm{stereo}$, which, in turn, is deemed more accurate than $d_\mathrm{apparent}$.
However, the availability of \ac{uwb} and stereo measurements may be inconsistent due to factors such as limited range, absence of the \ac{uwb} beacon or stereo camera, radio interference, and camera blur.
To address these limitations and utilize the most reliable data available, we select the best distance measurement and adjust the covariance matrix $\bm{\Sigma}$ accordingly:

\begin{align}
  \mathbf{z} &= {^C}\mathbf{R} \left( {^C}d \vec{\mathbf{d}} \right) + {^C}\mathbf{p}, \\
  \left\{ {^C}d,~ \bm{\Sigma} \right\} &= \begin{cases}
    \left\{ d_\mathrm{UWB},~ {^C}\mathbf{R} \bm{\Sigma}_\mathrm{UWB} {^C}\mathbf{R}^\top \right\}, \text{ if } d_\mathrm{UWB} \text{ available,} \\
    \left\{ d_\mathrm{stereo},~ {^C}\mathbf{R} \bm{\Sigma}_\mathrm{stereo} {^C}\mathbf{R}^\top \right\}, \begingroup \renewcommand{\arraystretch}{0.9} \begin{array}{l}\text{if } d_{\text{stereo}} \text{ avail. \rev{and}}\\ d_{\text{\rev{UWB}}} \text{\rev{ not available,}}\end{array} \endgroup \\
    \left\{ d_\mathrm{apparent},~ {^C}\mathbf{R} \bm{\Sigma}_\mathrm{apparent} {^C}\mathbf{R}^\top \right\}, \text{ otherwise}.
  \end{cases}
\end{align}

Here, ${^C}\mathbf{R} \in \mathbb{R}^{3 \times 3}$ and ${^C}\mathbf{p}=[{^C}p_x, {^C}p_y, {^C}p_z]^\top$ represent the camera's rotation matrix and position, respectively, describing the camera's pose in the world frame.
The covariance matrices for each type of distance measurement are defined as:
\vspace*{-0.6em}
\begin{align}
  \bm{\Sigma}_\mathrm{UWB} &= \diag\left( \sigma_{xy}^2, \sigma_{xy}^2, \sigma_{z, \mathrm{UWB}}^2 \right), \\
  \bm{\Sigma}_\mathrm{stereo} &= \diag\left( \sigma_{xy}^2, \sigma_{xy}^2, \sigma_{z, \mathrm{stereo}}^2 \right), \\
  \bm{\Sigma}_\mathrm{apparent} &= \diag\left( \sigma_{xy}^2, \sigma_{xy}^2, \sigma_{z, \mathrm{apparent}}^2 \right),
\end{align}
where $\sigma_{xy}$ represents the uncertainty in determining the bounding box's center, and $\sigma_{z, \mathrm{UWB}}$, $\sigma_{z, \mathrm{stereo}}$, $\sigma_{z, \mathrm{apparent}}$ reflect the uncertainties associated with the respective distance estimation methods.
These uncertainties are either empirically determined or based on the known characteristics of the sensors used.
It is assumed that the camera's optical axis aligns with the $z$-axis in the camera frame.

%%}

%%% END SECTION ============================================================

%%% START SECTION ==========================================================

%%{ SECTION: Formation control

\subsection{Formation control}
\label{sec:formation_control}

The proposed formation control strategy combines a leader-follower approach with receding horizon control. 
One \ac{uav} acts as the leader, equipped with onboard sensors and modules to detect the human worker and recognize gestures. The information obtained by the leader is shared with other \acp{uav} (see~\autoref{fig:softwareArchitecture}), which use it to maintain a desired formation relative to the worker's position. 
All \acp{uav} within the formation keep their cameras oriented towards the worker, thus, capturing additional perspectives to enhance the safety monitoring. This setup provides a diverse and comprehensive view of the monitored scene.
The state of the $i$-th~\ac{uav} is defined as ${^i}\mathbf{x} = [{^i}\mathbf{p}, {^i}\varphi, {^i}\xi]^\top \in \mathbb{R}^5$, where ${^i}\mathbf{p}=[{^i}p_x, {^i}p_y, {^i}p_z]^\top \in \mathbb{R}^3$ is the \ac{uav}'s position, ${^i}\varphi$, ${^i}\xi$ are its camera heading and pitch angles, respectively. The index $i=L$ refers to the leader~\ac{uav}, and $i \in \mathbb{N}_{>0}$ refer to the follower~\acp{uav}. The human worker's state is denoted as ${^H}\mathbf{x} = [{^H}\mathbf{p}, {^H}\varphi, 0]^\top$.

\begin{figure}[bt]
  \centering
  % left - bottom - right - top
  \adjincludegraphics[trim={{.0\width} {.095\height} {.0\width} {.03\height}}, clip, width=0.73\columnwidth]{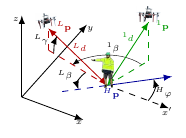}
  
  \vspace{-0.2em}
  \caption{Illustration of the proposed formation scheme for tracking a human worker, providing a diverse view of the scene from multiple angles (${^i}\beta$ and ${^i}\gamma$) and varying distances (${^i}d$).}
  \label{fig:formation_scheme}
  \vspace{-1.1em}
\end{figure}

To allow dynamic input from both operators and monitored workers, we introduce adaptive parameters\rev{--- desired observation angles ${^i}\beta$, ${^i}\gamma$, and distances ${^i}d$ (see~\autoref{fig:formation_scheme}).}
These parameters adapt based on gestures made by the human worker and requests communicated by the remote operator, which enables mid-flight adaptation of perspectives while continuously tracking human workers and interacting with them.
Each command, issued by detected gesture or operator's command, is mapped to an incremental change in a specific parameter (or a set of parameters) implemented as $q(k) = q(k-1) + i_q$, $i_q \in \mathbb{R}$. 
Parameter changes take effect at the next trajectory optimization step and typically result in observable \ac{uav} behavior within 1--2 seconds, depending on system latency and dynamics.
The incremental adaptation of formation parameters in response to gestures enhances the workers' situational awareness by allowing them to observe the \acp{uav}' behavior while trajectory generation process is designed to accommodate these incremental changes, resulting in smooth and feasible trajectories. 
Depending on the choice of parameters, the request results in a change of relative position of the entire formation to the monitored worker or incorporates also adaptation of mutual positions of individual \acp{uav} to reach desired set of perspectives.

Given the desired observation angles ${^L}\beta$ and ${^L}\gamma$, and distances ${^L}d$ to human worker, the leader's desired state is:
\begin{equation}\label{eq:leader_ref}
  {^L}\mathbf{x} = [{^H}\mathbf{p}^\top, \mathbf{0}]^\top - \left[\begin{array}{c}  {^L}d\cos({^H}\varphi-{^L}\beta)\cos({^L}\gamma) \\ {^L}d\sin({^H}\varphi-{^L}\beta)\cos({^L}\gamma) \\ {^L}d\sin(-{^L}\gamma) \\ {^L}\beta - {^H}\varphi \\ {^L}\gamma \end{array}\right].
\end{equation}
Similarly, the desired state of follower~\acp{uav}, given the observation distance ${^i}d$ and angles ${^i}\beta$ and ${^i}\gamma$ defined relative to the leader~\ac{uav}'s observation angles, is computed as:
\begin{equation}\label{eq:follower_ref}
  {^i}\mathbf{x} = [{^H}\mathbf{p}^\top, \mathbf{0}]^\top - \left[\begin{array}{c}  {^i}d\cos({^L}\varphi-{^i}\beta)\cos({^i}\gamma-{^L}\xi) \\ {^i}d\sin({^L}\varphi-{^i}\beta)\cos({^i}\gamma-{^L}\xi) \\ {^i}d\sin({^L}\xi-{^i}\gamma) \\ {^i}\beta-{^L}\varphi \\ {^i}\gamma - {^L}\xi \end{array}\right]. 
\end{equation}
Here, ${^L}\beta = 0$ represents an observation angle aligned with the heading of the worker ${^H}\varphi$. \rev{Note that ${^H}\varphi$ does not necessarily match the orientation of the worker's body; it can be also set to coincide with the estimated motion direction or be set to a constant value.}

The formation controller first applies~\eqref{eq:leader_ref} and~\eqref{eq:follower_ref} to every pose on the prediction horizon, using the worker's predicted trajectory and the leader's planned trajectory. \rev{Initially, reference trajectories are generated without considering collisions to reduce computational load. Then, safe paths are planned along reference trajectories using environmental maps, followed by trajectory optimization within safe corridors generated using convex decomposition of free space\cite{Liu2017}. To ensure inter-UAV safety, planned trajectories (inflated by a safety margin $\Gamma_\mathrm{dis}$) are treated as dynamic obstacles for other \acp{uav}. This three-stage planning process runs onboard, enabling real-time adaptation to dynamic environmental changes and requests for view adaptation.}
For a detailed explanation of the \ac{uav} coordination method, refer to~\cite{kratky2021aerialfilming}.

%%}

%%% END SECTION ============================================================

%%% START SECTION ==========================================================

%%{ SECTION: Results

%%{ SECTION: Aerial Platform

\section{Results}
\label{sec:experimental_results}

The proposed \ac{hsi} approach was evaluated through both Gazebo simulations and field experiments conducted in a mock-up scenario. The simulations were performed using the MRS software stack~\cite{Baca2020mrs} on a computer with an i7-10510U processor and 16GB of RAM. Videos of \rev{the simulations and experiments} can be found at \url{https://mrs.felk.cvut.cz/gestures2024}, with example frames shown in~\autoref{fig:system_demo}.
\rev{The value $^H\varphi = 0$ is used in all presented simulations and experiments.  }

%%% END SECTION ============================================================

%%% START SECTION ==========================================================

\subsection{Simulation}
\label{sec:simulation}

\begin{figure}[bt]
  \centering

  \vspace*{0.06cm}
  \begin{tikzpicture}
    \node[anchor=south west,inner sep=0] (a) at (0,0) {
      \adjincludegraphics[trim={{.0\width} {.00\height} {.11\width} 
      {.00\height}},clip,width=\columnwidth]{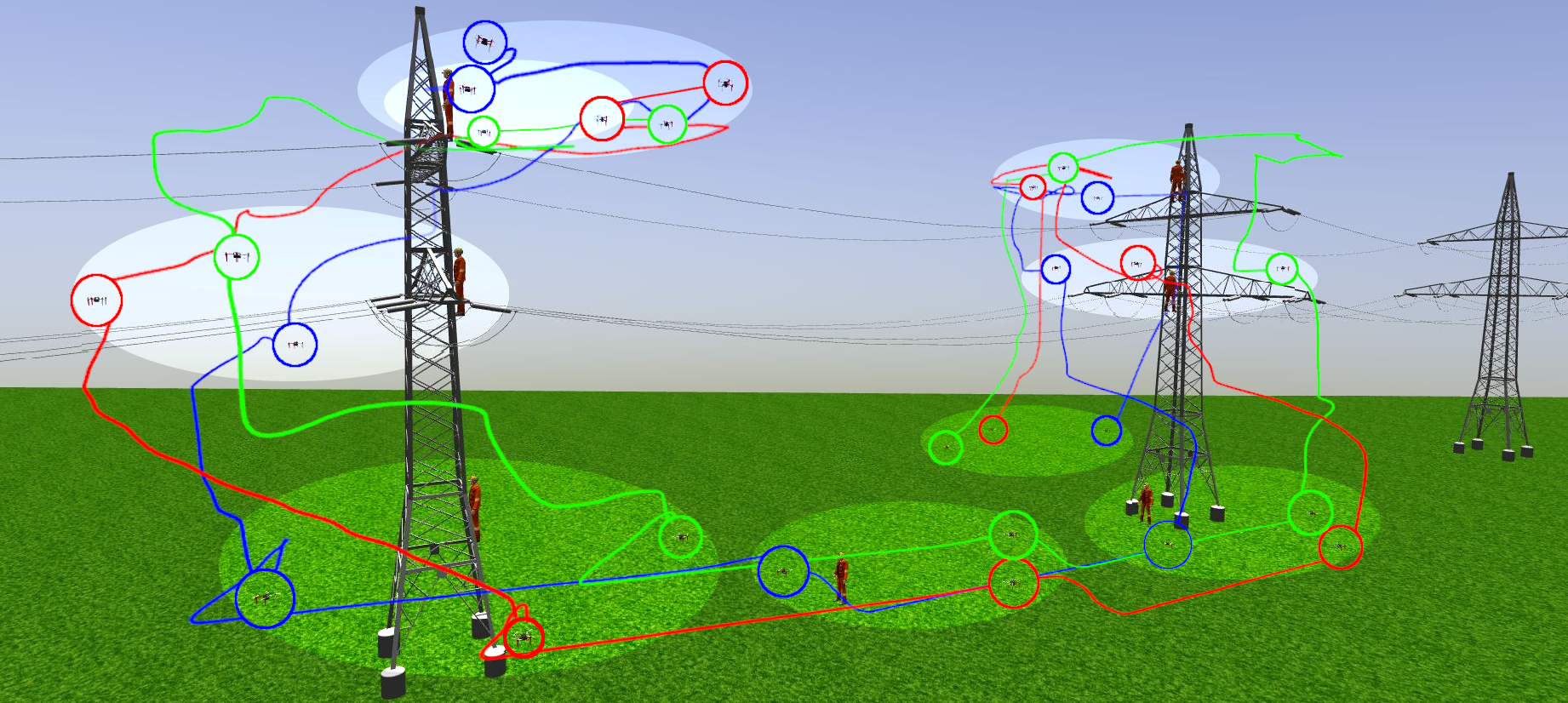}
    };

    \begin{scope}[x={(a.south east)},y={(a.north west)}]

    %%{ useful grid to help you find coordinates for plotting the overlay
    % \draw[black, xstep=.1, ystep=.1] (0,0) grid (1,1);
    % \foreach \i in {0,0.1,0.2,0.3,0.4,0.5,0.6,0.7,0.8,0.9,1} {
    %   \node[align=center] at (\i, -0.05) {\i};
    %   \node[align=center] at (\i, 1.05) {\i};
    %   \node[align=center] at (-0.05, \i) {\i};
    %   \node[align=center] at (1.05, \i) {\i};
    % }
    %%}

        \node[fill=black, fill opacity=0.0, text=white, text opacity=1.0] at (0.63, 0.382) {\footnotesize \textbf{0 s}};
        \node[fill=black, fill opacity=0.0, text=black, text opacity=1.0] at (0.790, 0.83) {\footnotesize \textbf{20 s}};
        \node[fill=black, fill opacity=0.0, text=black, text opacity=1.0] at (0.960, 0.66) {\footnotesize \textbf{52 s}};
        \node[fill=black, fill opacity=0.0, text=white, text opacity=1.0] at (0.94, 0.15) {\footnotesize \textbf{77 s}};
        \node[fill=black, fill opacity=0.0, text=white, text opacity=1.0] at (0.66, 0.075) {\footnotesize \textbf{98 s}};
        \node[fill=black, fill opacity=0.0, text=white, text opacity=1.0] at (0.17, 0.07) {\footnotesize \textbf{121 s}};
        \node[fill=black, fill opacity=0.0, text=black, text opacity=1.0] at (0.05, 0.68) {\footnotesize \textbf{137 s}};
        \node[fill=black, fill opacity=0.0, text=black, text opacity=1.0] at (0.21, 0.90) {\footnotesize \textbf{172 s}};
        \node[fill=black, fill opacity=0.0, text=black, text opacity=1.0] at (0.575, 0.92) {\footnotesize \textbf{182 s}};

        \draw [->] (0.25, 0.88) -- (0.276, 0.854);

    \end{scope}

  \end{tikzpicture}

  \vspace{-0.3em}
  \caption{Simulation emulating a safety monitoring scenario where three \acp{uav} respond to multiple view adaptation requests. Ellipses indicate the \ac{uav} formation at specific time instances, while the trajectories are represented by colored lines: red for the leader \ac{uav}, and blue and green for the follower \acp{uav}.}
  \label{fig:simulation}
\vspace{-0.9em}
\end{figure}
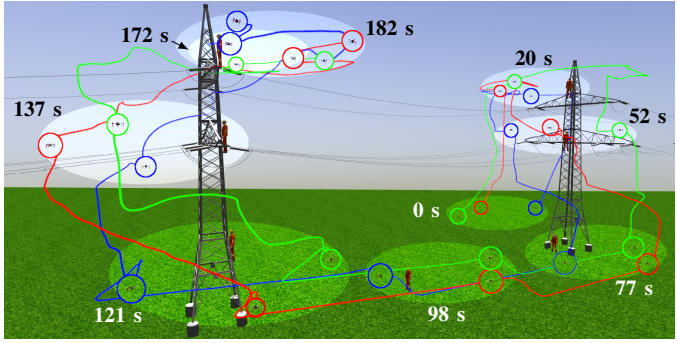

In the simulation scenarios, designed to closely replicate real-world applications of the proposed methodology, we focus on the evaluation of the planning and formation control part of the system while simulating the rest of the system as simplified modules providing data with predefined uncertainty.
In one scenario, a formation of three \acp{uav} is tasked with monitoring a human worker performing maintenance operations at two power transmission towers (see~\autoref{fig:simulation}).
Throughout the mission, the formation received 25 requests to adjust the views provided by the \acp{uav}. 
The requests contain modifying both the observation angles and the distance of all \acp{uav} to the worker's estimated position simultaneously, and modifications of these parameters only for a subset of \acp{uav}.
Hence, the applied changes of parameters represent both shifts in a relative position and orientation of the whole formation to the worker and changes of parameters leading to adaptation of the formation shape.

\begin{figure}[bt]
  \centering
  \adjincludegraphics[trim={{.02\width} {.03\height} {.0\width} {.020\height}}, clip, width=1.0\columnwidth]{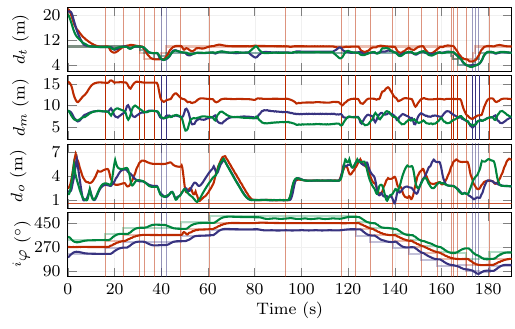}
  \vspace*{-1.7em}
  \caption{Simulation timeline showing the evolution of the key metrics over time: the distances between \acp{uav} and the human worker ($d_t$), the\rev{ distances between individual pairs of} \acp{uav} ($d_m$), the distance to the nearest obstacle ($d_o$), and the observation angle \rev{(${^i}\varphi$)}. These metrics are displayed separately for each \ac{uav}, with the leader's data in red and the followers' data in green and blue \rev{(except for $d_m$ where green and blue represents distance between leader and individual followers, and red represents distance between followers)}. The transparent lines represent the reference values for individual quantities. Vertical lines indicate the moments when commands were received from the operator (red) and the human worker (blue).}
  \label{fig:simulation_timeline}
  \vspace{-0.8em}
\end{figure}

Most of the requests were initiated by a remote operator monitoring safety compliance, while the remaining requests were triggered by commands from the human worker. The worker's commands aimed to increase the \acp{uav}' relative distance to ensure comfort and safety due to wind conditions and proximity to obstacles.
The mission demonstrated the system's capability to navigate safely near obstacles, including maneuvering through narrow gaps formed by electrical power lines while continuously tracking the human subject. As illustrated in~\autoref{fig:simulation_timeline}, the \acp{uav} successfully maintained the required distances from both the target and surrounding obstacles, while adhering to the desired observation angles throughout the mission. 

%%% END SECTION ============================================================

%%% START SECTION ==========================================================

\subsection{Real-world experiments}
\label{sec:real_world_experiments}

The integration of the introduced modules into a unified system running onboard real \acp{uav} was demonstrated through field experiments involving three~\acp{uav} collaborating with a human worker.
The worker's gestures were mapped to changes in formation parameters to demonstrate real-world responsiveness in the following way: crossing arms (gesture $\text{ID} = 1$) decreased ${^L}\beta$, extending an arm to the side ($\text{ID} = 2$) increased ${^L}\beta$, placing palms together ($\text{ID} = 3$) decreased ${^L}\gamma$, and raising an arm upward ($\text{ID} = 4$) increased ${^L}\gamma$. The increments and decrements for ${^L}\beta$ and ${^L}\gamma$ were set to \SI{30}{\degree} and \SI{5}{\degree}, respectively. The human worker's heading, ${^H}\varphi$, was assumed to be constant, and gestures were filtered using the twenty most recent measurements. Additional parameters used in the experiments are listed in~\autoref{tab:tableParamters}.

\begin{table}[tb]
    \setlength{\tabcolsep}{3pt}
    \centering
    \vspace*{0.14cm}
    \caption{Values of parameters used in the experiments. Some of these parameters are influenced by the safety monitoring requirements of the industrial partners in the AERIAL-CORE project.}
    \vspace*{-0.1cm}
    \begin{adjustbox}{max width=1.00\columnwidth}
	\begin{tabular}{lcc|lcc}
		  \hline
		  \textbf{Parameter} & \textbf{Symbol} & \textbf{Value} & \textbf{Parameter} & \textbf{Symbol} & \textbf{Value}\\
		  \hline
  \ac{cnn} sliding window & N & \SI{9}{[-]}                                           & $^L$\ac{uav} obs. heading & ${^L}\beta$ & \SI{90}{\degree}\\
  Data filtering thr. & $t_c$ & \SI{20}{\second}                                      & $^L$\ac{uav} obs. pitch & ${^L}\gamma$ & \SI{11}{\degree} \\
  Ratio threshold & $\Pi_d$ & \SI{0.8}{[-]}                                           & $^L$\ac{uav} des. distance & ${^L}d$ & \SI{10.00}{\meter} \\
  Command threshold & $t_d$ & \SI{5}{\second}                                         & $^1$\ac{uav} obs. heading & ${^1}\beta$ & \SI{60}{\degree} \\
  Pos. process noise  & $\sigma_{p_\bullet}$ & $ \SI{0.1}{\metre} $                      & $^1$\ac{uav} obs. pitch & ${^1}\gamma$ & \SI{0}{\degree} \\
  Vel. process noise  & $\sigma_{v_\bullet}$ & $ \SI{0.1}{\metre \per \second} $         & $^1$\ac{uav} des. distance & ${^1}d$ & \SI{8.00}{\meter}  \\
  Direction meas. noise & $\sigma_{xy}$ & $ \SI{0.05}{\metre} $                       & $^2$\ac{uav} obs. heading & ${^2}\beta$ & \SI{-60}{\degree} \\
  UWB meas. noise     & $\sigma_{z,\mathrm{UWB}}$ & $ \SI{0.1}{\metre} $                & $^2$\ac{uav} obs. pitch & ${^2}\gamma$ & \SI{0}{\degree} \\
  Stereo meas. noise     & $\sigma_{z,\mathrm{stereo}}$ & $ \SI{0.3}{\metre} $          & $^2$\ac{uav} des. distance & ${^2}d$ & \SI{8.00}{\meter} \\
  Apparent size meas. noise     & $\sigma_{z,\mathrm{apparent}}$ & $ \SI{0.6}{\metre} $ & Mutual distance thr. & $\Gamma_\mathrm{dis}$ & \SI{2.50}{\meter} \\
		  \hline
		\end{tabular}
    \end{adjustbox}
    \vspace*{-0.6em}
    \label{tab:tableParamters}
\end{table}

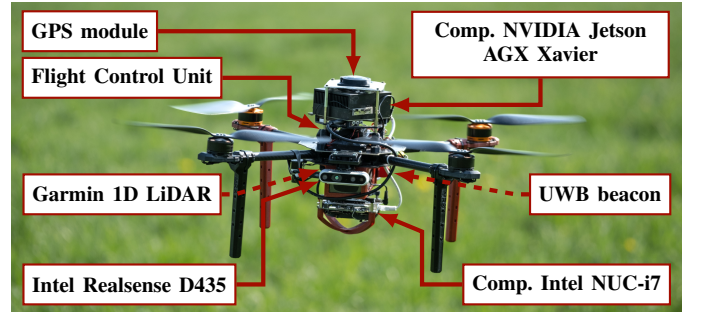
\begin{figure}
  \centering
  \begin{tikzpicture}[|every node/.style={
  inner sep = .01em,
  outer sep=0.02em,
  text centeblue,
  minimum height = .5em}]

  \definecolor{color_blue}{HTML}{3321E0}
  \definecolor{color_red}{HTML}{A30D00}
  \definecolor{color_green}{rgb}{0, .522, .243}

\pgfmathsetmacro{\arrowshift}{0.0003em} % shift in vertical direction

  \node[anchor=south west,inner sep=0] (a) at (0,0) {
      \def\arraystretch{0}%
      {\adjincludegraphics[trim={{.02\width} {.16\height} {.00\width} 
      {.16\height}},clip,width=\columnwidth]{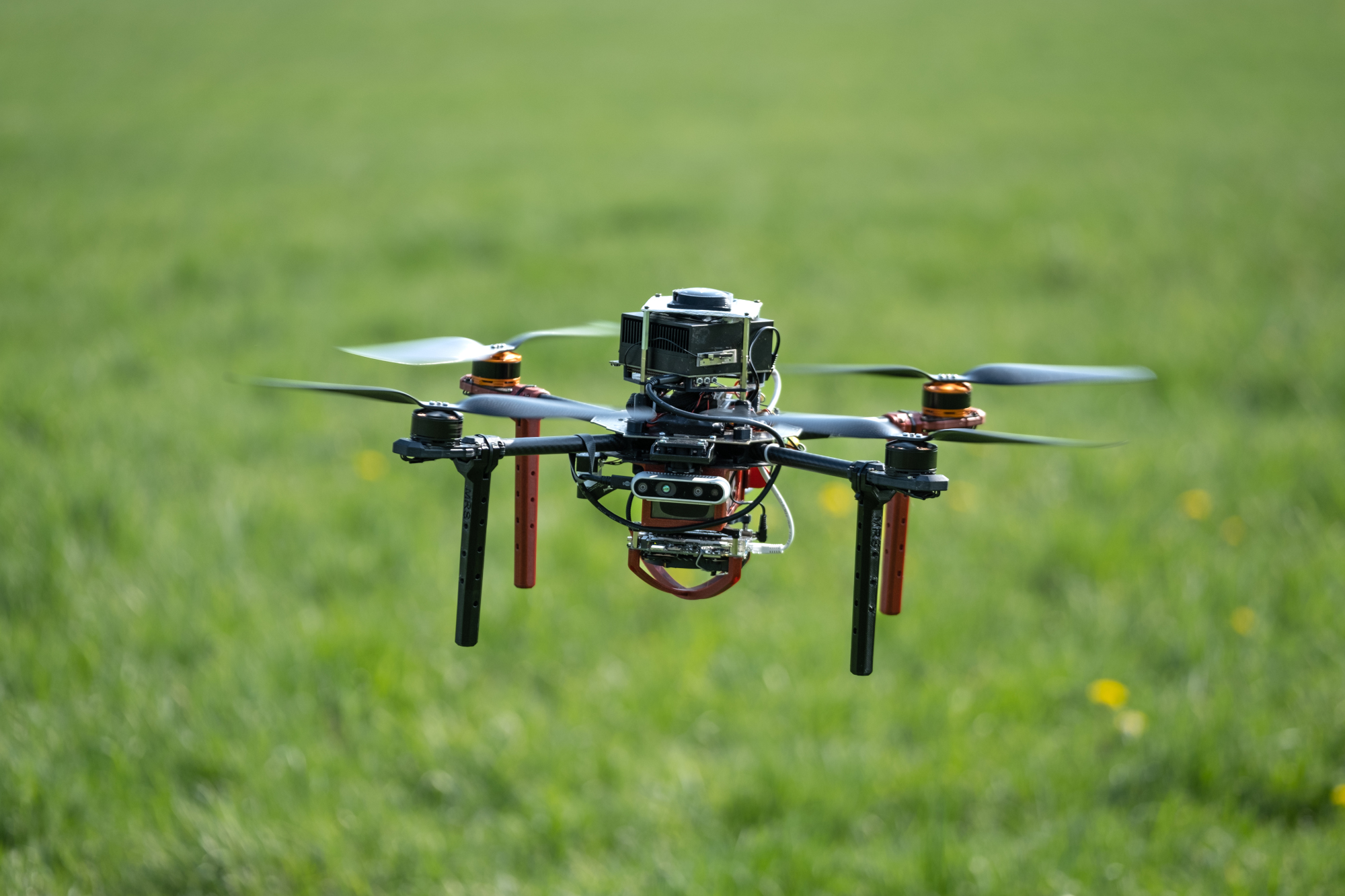}}
      }; 
  \begin{scope}[x={(a.south east)},y={(a.north west)},font=\footnotesize]

    % { grid useful grid to help you find coordinates for plotting the overlay
     % \draw[black, xstep=.1, ystep=.1] (0,0) grid (1,1);
     % \foreach \i in {0,0.1,0.2,0.3,0.4,0.5,0.6,0.7,0.8,0.9,1} {
     %   \node[align=center] at (\i, -0.05) {\i};
     %   \node[align=center] at (\i, 1.05) {\i};
     %   \node[align=center] at (-0.05, \i) {\i};
     %   \node[align=center] at (1.05, \i) {\i};
     % }
    % }

    \coordinate (realsense) at (0.176,0.09);
    \coordinate (leds) at (0.333,0.18);
    \coordinate (uwb) at (0.878,0.38);
    \coordinate (xavier) at (0.789,0.862);
    \coordinate (nuc) at (0.828,0.09);
    \coordinate (pixhawk) at (0.165,0.755);
    \coordinate (gps_module) at (0.117,0.905);
    \coordinate (garmin) at (0.165,0.38);
    
    \coordinate (realsense_t) at (0.46,0.422);
    \coordinate (leds_t) at (0.333,0.390);
    \coordinate (leds_t2) at (0.246,0.18);
    \coordinate (uwb_t) at (0.57,0.448);
    \coordinate (xavier_t) at (0.563,0.668);
    \coordinate (nuc_t) at (0.545,0.32);
    \coordinate (pixhawk_t) at (0.46,0.59);
    \coordinate (gps_module_t) at (0.51,0.76);
    \coordinate (garmin_t) at (0.46,0.46);

    \node[draw=color_red, fill=white, minimum width=25pt, minimum height=15pt, thick] (realsense_label) at (realsense) {\textcolor{black}{\textbf{Intel Realsense D435}}};
    \node[draw=color_red, fill=white, minimum width=25pt, minimum height=15pt, thick] (garmin_label) at (garmin) {\textcolor{black}{\textbf{Garmin 1D LiDAR}}};
    \node[draw=color_red, fill=white, minimum width=25pt, minimum height=15pt, thick] (uwb_label) at (uwb) {\textcolor{black}{\textbf{UWB beacon}}};
    \node[draw=color_red, minimum width=25pt, fill =white, minimum height=15pt, thick] (pixhawk_label) at (pixhawk) {\textcolor{black}{\textbf{Flight Control Unit}}};
    \node[draw=color_red, fill=white, minimum width=25pt, minimum height=15pt, thick] (nuc_label) at (nuc) {\textcolor{black}{\textbf{Comp. Intel NUC-i7}}};
    \node[draw=color_red, fill=white, minimum width=25pt, minimum height=15pt, thick] (gps_module_label) at (gps_module) {\textcolor{black}{\textbf{GPS module}}};
    \node[draw=color_red, fill=white, minimum width=25pt, minimum height=15pt, thick] (xavier_label) at (xavier) {\textcolor{black}{$\begin{array}{c} \textbf{Comp. NVIDIA Jetson} \\ \textbf{AGX Xavier} \end{array}$}};

     \draw [->, ultra thick, color = color_red] (gps_module_label.east) -| (gps_module_t);
     \draw [->, ultra thick, color = color_red] (xavier_label.south) |- (xavier_t);
     \draw [->, ultra thick, color = color_red] (nuc_label.west) -| ($(nuc_label.west) + (-0.06, 0.0)$) |- ($(nuc_label.west) + (-0.06, 0.17)$) --(nuc_t);
     \draw [->, ultra thick, color = color_red] (pixhawk_label.east) -| ($(pixhawk_label.east) + (+0.1, 0.0)$) |- ($(pixhawk_label.east) + (+0.10, -0.12)$) -- (pixhawk_t);
     \draw [->, ultra thick, color = color_red] (realsense_label.east) -| ($(realsense_label.east) + (+0.04, 0.0)$) |- ($(realsense_label.east) + (+0.04, 0.27)$) -- (realsense_t);
     \draw [->, dashed, ultra thick, color = color_red] (uwb_label.west) -| ($(uwb_label.west) + (-0.02, 0.0)$) -- (uwb_t);
     \draw [->, dashed, ultra thick, color = color_red] (garmin_label.east) -| ($(garmin_label.east) + (0.02, 0.0)$) -- (garmin_t);
     % \draw [->, ultra thick, color = color_red] (leds_label.west) -- (leds_t2);

  \end{scope}

\end{tikzpicture} 
 \vspace*{-1.5em}
 \caption{The primary aerial platform for the experiments.}
  \label{fig:uav_platform}
  \vspace{-1.6em}
\end{figure}

Two types of multi-rotor \acp{uav} were used in the experimental validation.
The primary \ac{uav} utilizes a Tarot 650 frame and is equipped with a Pixhawk \ac{fcu} with sensors for \ac{uav} state estimation, gesture recognition, and human detection (see~\autoref{fig:uav_platform}).
% ---  RGBD camera Intel\textsuperscript{\tiny\textregistered} Realsense D435, Terabee UWB beacon, 1D LiDAR Garmin LiDAR-Lite v3, and 9-DOF \ac{imu} integrated in Pixhawk \ac{fcu}. 
Onboard computation is handled by an NVIDIA Jetson AGX Xavier computer for human detection and gesture recognition, while an Intel NUC-i7 is used for state estimation, control, and planning tasks.
The computers are interconnected via an Ethernet interface, ensuring reliable data transfer.
While it is feasible to run the entire pipeline on a single AGX Xavier computer, utilizing additional computational resources allows for faster image processing and segregates the computationally intensive image processing pipeline from the safety-critical modules necessary for autonomous \ac{uav} flight.
The secondary \acp{uav} are constructed using F450 platforms with a payload limited to a single onboard computer Intel NUC-i7, Pixhawk \ac{fcu}, and the essential sensors for state estimation and scene capture. The \acp{uav} fuse information from GNSS, IMU and 1D LiDAR for self-state estimation. A detailed description of the platforms can be found in~\cite{HertJINTHWpaper, hert2022MRSModularUAV}.

The final evaluation of the system was conducted through a series of experiments involving different numbers of \acp{uav} and varied environments to fine-tune the performance of individual modules and their integration.
The presented evaluation is based on three autonomous flights conducted under a consistent setup and module configuration. During these flights, the system achieved a success rate of $87\%$ in propagating human gestures to scene view adaptations.
\rev{A propagation of gesture to shape adaptation is considered successful if it is executed while the human is performing the gesture.}

\begin{figure}[tb]
  \centering
  \vspace*{0.10cm}
  \adjincludegraphics[trim={{.02\width} {.028\height} {.0\width} {.028\height}}, clip, width=1.0\columnwidth]{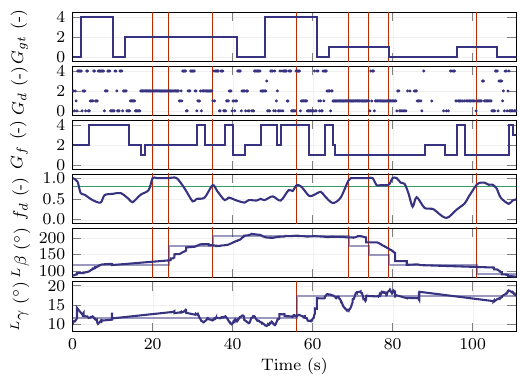}
  \vspace*{-1.6em}
  \caption{A timeline illustrating the process from human gestures to the adaptation of the relative view provided by the \acp{uav}. The graphs display the IDs of gestures performed by the human worker ($G_{gt}$), gestures detected by the gesture recognition module ($G_d$), the dominant gesture ($G_f$) and its relative frequency ($f_d$), and the observation angles of the leading \ac{uav} ($^L\beta$, $^L\gamma$\rev{, including reference angles shown as transparent lines}).
  Each gesture is represented by its associated ID (see~\autoref{sec:experimental_results}). Gesture $\text{ID} = 0$ indicates that the human is not performing any gesture. 
  The green line represents the threshold $\Pi_d$ on $f_d$, while the vertical red lines mark instances of confirmed requests for scene view adaptation.}
  \label{fig:timeline}
  \vspace*{-0.5cm}
\end{figure}

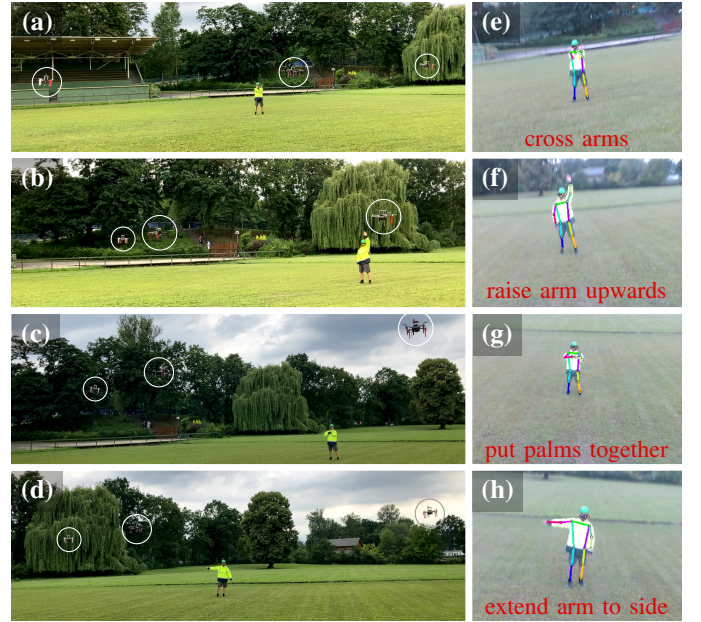
\begin{figure}[tb]
  \centering
  \vspace*{0.1cm}
  \begin{tikzpicture}
    \node[anchor=south west,inner sep=0] (a) at (0,0) {
        \includegraphics[width=1.0\columnwidth]{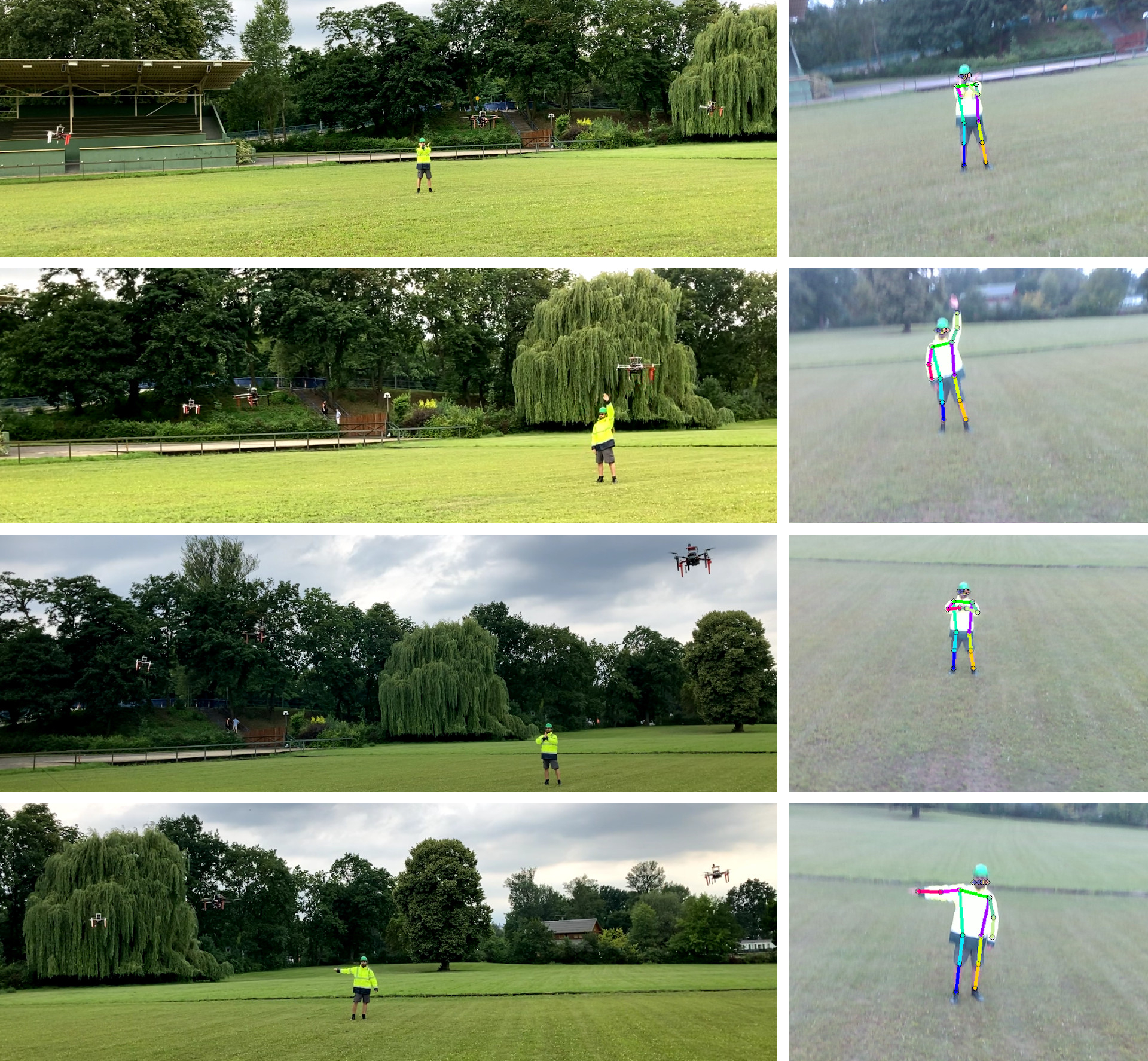}
    };

    \pgfmathsetmacro{\hzero}{0.039}
    \pgfmathsetmacro{\vzero}{0.9658}
    \pgfmathsetmacro{\hshift}{0.6863}
    \pgfmathsetmacro{\vshift}{-0.2520}
    \pgfmathsetmacro{\hoffset}{0.00}
    \pgfmathsetmacro{\voffset}{-0.00}
    \pgfmathsetmacro{\vshiftr}{0.2515}
   
    \begin{scope}[x={(a.south east)},y={(a.north west)}]

    %%{ useful grid to help you find coordinates for plotting the overlay
    % \draw[black, xstep=.1, ystep=.1] (0,0) grid (1,1);
    % \foreach \i in {0,0.1,0.2,0.3,0.4,0.5,0.6,0.7,0.8,0.9,1} {
    %   \node[align=center] at (\i, -0.05) {\i};
    %   \node[align=center] at (\i, 1.05) {\i};
    %   \node[align=center] at (-0.05, \i) {\i};
    %   \node[align=center] at (1.05, \i) {\i};
    % }
    %%}

        \node[fill=black, fill opacity=0.3, text=white, text opacity=1.0] at (\hzero, \vzero) {\textbf{(a)}};
        \node[fill=black, fill opacity=0.3, text=white, text opacity=1.0] at (\hzero+\hshift-0.002, \vzero) {\textbf{(e)}};
        \node[fill=black, fill opacity=0.3, text=white, text opacity=1.0] at (\hzero, \vzero+\vshift) {\textbf{(b)}};
        \node[fill=black, fill opacity=0.3, text=white, text opacity=1.0] at (\hzero+\hshift-0.004, \vzero+\vshift-0.0008) {\textbf{(f)}};
        \node[fill=black, fill opacity=0.3, text=white, text opacity=1.0] at (\hzero-0.0023, \vzero+2*\vshift +0.0005) {\textbf{(c)}};
        \node[fill=black, fill opacity=0.3, text=white, text opacity=1.0] at (\hzero+\hshift-0.0008, \vzero+2*\vshift) {\textbf{(g)}};
        \node[fill=black, fill opacity=0.3, text=white, text opacity=1.0] at (\hzero, \vzero+3*\vshift+\voffset) {\textbf{(d)}};
        \node[fill=black, fill opacity=0.3, text=white, text opacity=1.0] at (\hzero+\hshift, \vzero+3*\vshift+\voffset) {\textbf{(h)}};

        \node[text=red!90!black, text opacity=1.0] at (0.843, 0.530) {\small raise arm upwards};
        \node[text=red!90!black, text opacity=1.0] at (0.843, 0.775) {\small cross arms};
        \node[text=red!90!black, text opacity=1.0] at (0.843, 0.275) {\small put palms together};
        \node[text=red!90!black, text opacity=1.0] at (0.843, 0.025) {\small extend arm to side};

  % UAV highlights
  \draw[white] (0.053,0.870) circle (5.5pt);
  \draw[white] (0.421,0.886) circle (5.5pt);
  \draw[white] (0.62,0.896) circle (4.5pt);

  \draw[white] (0.1666,0.617) circle (4.5pt);
  \draw[white] (0.221,0.627) circle (6.5pt);
  \draw[white] (0.555,0.653) circle (6.5pt);

  \draw[white] (0.603,0.473) circle (6.5pt);
  \draw[white] (0.125,0.375) circle (4.5pt);
  \draw[white] (0.22,0.402) circle (5.5pt);

  \draw[white] (0.086,0.132) circle (4.5pt);
  \draw[white] (0.187,0.149) circle (5.5pt);
  \draw[white!60!black] (0.624,0.174) circle (5.5pt);

    \end{scope}

  \end{tikzpicture}
  \vspace*{-1.5em}
  \caption{A sequence of example frames showing a team of \acp{uav} following a human worker (a)-(d) and adapting the relative view based on detected gestures (e)-(h). The experiment demonstrates a full 3D deployment, requiring adjustments to the observation angles in both horizontal and vertical directions.}
  \vspace*{-1.4em}
  \label{fig:system_demo}
\end{figure}

A timeline illustrating the propagation of a human gesture through the gesture recognition and filtering pipeline to the adaptation of observation angles during one of the experiments is shown in~\autoref{fig:timeline}.
The average time from the initiation of a gesture to the onset of shape adaptation during the final experiments was \SI{7}{\second}. %The average time from the initiation of a gesture to the onset of shape adaptation during the final experiments was \SI{7}{\second}. 
This response time is influenced by a conservative parameter setup, which is necessary to prevent undesired view adaptations caused by incorrect gesture classification or a temporary worker's pose resembling one of the predefined gestures.
Throughout the experiments, the \ac{uav} team effectively maintained the safety distance $\Gamma_\mathrm{dis}$ between the human and among the \acp{uav}, while keeping their cameras oriented toward the worker. This ensured safety and demonstrated the capability of the proposed approach in real-world scenarios (see \autoref{fig:system_demo}).

%%}

%%% END SECTION ============================================================

%%% START SECTION ==========================================================

%%{ SECTION: Discussion

\section{Discussion}
\label{sec:discussion}

The conducted experiments underscore the potential of using hand gestures for intuitive control and coordination of multi-robot aerial systems.
This approach is particularly beneficial in scenarios such as safety monitoring and assisting human workers in challenging environments, as it does not impose additional workload on the workers or require extra equipment for conventional wireless communication.
However, gesture-based control presents specific challenges distinct from other modes of interaction. 

Firstly, the permissible observation angles and distance ranges are limited by the performance of the gesture recognition module (mainly effective recognition range) and the safety requirements of the workers, which must be considered in the process of adapting the scene view. 
Our approach addresses this challenge by imposing stringent limits on the parameters ${^L}\gamma$, ${^i}d$ and $\Gamma_\mathrm{dis}$.
In future work, we aim to enhance the system's robustness by implementing a worker detection pipeline across multiple \acp{uav}, coupled with distributed estimation of the worker's position~\cite{HoIROS2021, Tallamraju2019RAL}.
This approach not only provides multiple perspectives but also decrease the risk of the loss of the tracked worker due to occlusion or obstacle avoidance maneuvers.

A significant challenge of gesture-based control is its dependence on sufficient quality of sensory data for vision-based human detection and gesture recognition, which compromises the robustness of this approach under various environmental conditions, such as operations in reduced visibility.
Combining multiple modalities, such as thermal imaging or LiDAR data, within the human detection and gesture recognition pipeline can significantly increase the reliability of such systems.
However, this comes at the cost of increased \ac{uav} weight and size, which is undesirable in scenarios where \acp{uav} operate in close proximity to humans.

Another critical aspect of using gestures to interact with teams of aerial robots is the potential for workers involved in maintenance tasks to unintentionally assume positions that resemble predefined gestures. This issue is further complicated by challenging environments in which these tasks occur, where constrained mobility and the need to maintain uncomfortable postures are common. Given the potential for misinterpretation, it is essential to configure the gesture processing pipeline carefully to prevent false positive detections from being translated into mission-related commands.

Lastly, our experimental campaigns have shown that providing clear feedback from the UAV formation to the human executing gestures is one of the most significant and often overlooked aspects of HSI via gestures.
To avoid the need for additional equipment for visual feedback, we have structured the behavior of the \acp{uav} so that the human can infer the acceptance of their commands based on the observable actions of the \acp{uav}. 
In this context, making incremental adjustments to the formation parameters and avoiding continuous scene view adaptation commands have proven beneficial.

%%}

%%% END SECTION ============================================================

%%% START SECTION ==========================================================

%%{ SECTION: Conclusion

\section{Conclusion}
\label{sec:conclusion}

In this paper, we introduced a novel approach for contactless~\acl{hsi} using hand gestures to control a team of~\acp{uav} in safety monitoring scenarios. The proposed approach enables safe and efficient interaction between remote human operators, human workers, and autonomous aerial systems, providing significant benefits in real-world applications. By integrating hand gestures as a control modality, human workers can command and adjust various formation parameters, request immediate assistance, and initiate other mission-related commands.
The proposed approach incorporates robust algorithms for human worker detection and gesture recognition, ensuring accurate and prompt responses. Simulations and field experiments validated the effectiveness of the approach, demonstrating successful navigation in complex environments while providing the required perspectives controlled both through remote commands and based on the detected hand gestures.

%%}

% %%{ SECTION: Bibliography
\bibliographystyle{IEEEtran}
\bibliography{main.bib}
% %%}

\end{document}